\documentclass[10pt,twocolumn,letterpaper]{article}

\usepackage{cvpr}              

\usepackage{graphicx}
\usepackage{amsmath}
\usepackage{amssymb}
\usepackage{booktabs}
\usepackage{multirow}
\usepackage[accsupp]{axessibility}
\usepackage[pagebackref,breaklinks,colorlinks]{hyperref}
\usepackage[capitalize]{cleveref}
\crefname{section}{Sec.}{Secs.}
\Crefname{section}{Section}{Sections}
\Crefname{table}{Table}{Tables}
\crefname{table}{Tab.}{Tabs.}

\def\cvprPaperID{15058} 
\def\confName{CVPR}
\def\confYear{2024}

\begin{document}

\title{Improving Bird's Eye View Semantic Segmentation by Task Decomposition}

\author{
Tianhao Zhao$^{1}$\thanks{Equal contribution.} 
\quad Yongcan Chen$^{1}$\footnotemark[1] 
\quad Yu Wu$^{1}$ 
\quad Tianyang Liu$^{1}$ 
\quad Bo Du$^{1}$\thanks{Corresponding author.} \\ 
\quad Peilun Xiao$^{2}$ 
\quad Shi Qiu$^{2}$ 
\quad Hongda Yang$^{2}$ 
\quad Guozhen Li$^{2}$ 
\quad Yi Yang$^{2}$ 
\quad Yutian Lin$^{1}$\footnotemark[2]\\
$^1$ Institute of Artificial Intelligence, School of Computer Science, \\
Hubei Luojia Laboratory, Wuhan University, Wuhan, China \\
$^2$ Didi Chuxing, China \\
\normalsize{\texttt{\{happytianhao, chenyongcan, wuyucs, bodu, yutian.lin\}@whu.edu.cn}}
}

\maketitle

\begin{abstract}
Semantic segmentation in bird's eye view (BEV) plays a crucial role in autonomous driving. Previous methods usually follow an end-to-end pipeline, directly predicting the BEV segmentation map from monocular RGB inputs. However, the challenge arises when the RGB inputs and BEV targets from distinct perspectives, making the direct point-to-point predicting hard to optimize. In this paper, we decompose the original BEV segmentation task into two stages, namely BEV map reconstruction and RGB-BEV feature alignment. In the first stage, we train a BEV autoencoder to reconstruct the BEV segmentation maps given corrupted noisy latent representation, which urges the decoder to learn fundamental knowledge of typical BEV patterns. The second stage involves mapping RGB input images into the BEV latent space of the first stage, directly optimizing the correlations between the two views at the feature level. Our approach simplifies the complexity of combining perception and generation into distinct steps, equipping the model to handle intricate and challenging scenes effectively. Besides, we propose to transform the BEV segmentation map from the Cartesian to the polar coordinate system to establish the column-wise correspondence between RGB images and BEV maps. Moreover, our method requires neither multi-scale features nor camera intrinsic parameters for depth estimation and saves computational overhead. Extensive experiments on nuScenes and Argoverse show the effectiveness and efficiency of our method. Code is available at \href{https://github.com/happytianhao/TaDe}{https://github.com/happytianhao/TaDe}.
\end{abstract}

\section{Introduction}
\begin{figure}[t]
    \centering
    \begin{minipage}{0.46\linewidth}
        \centering
        \subfloat[Perspective View RGB Image]{\includegraphics[width=\linewidth]{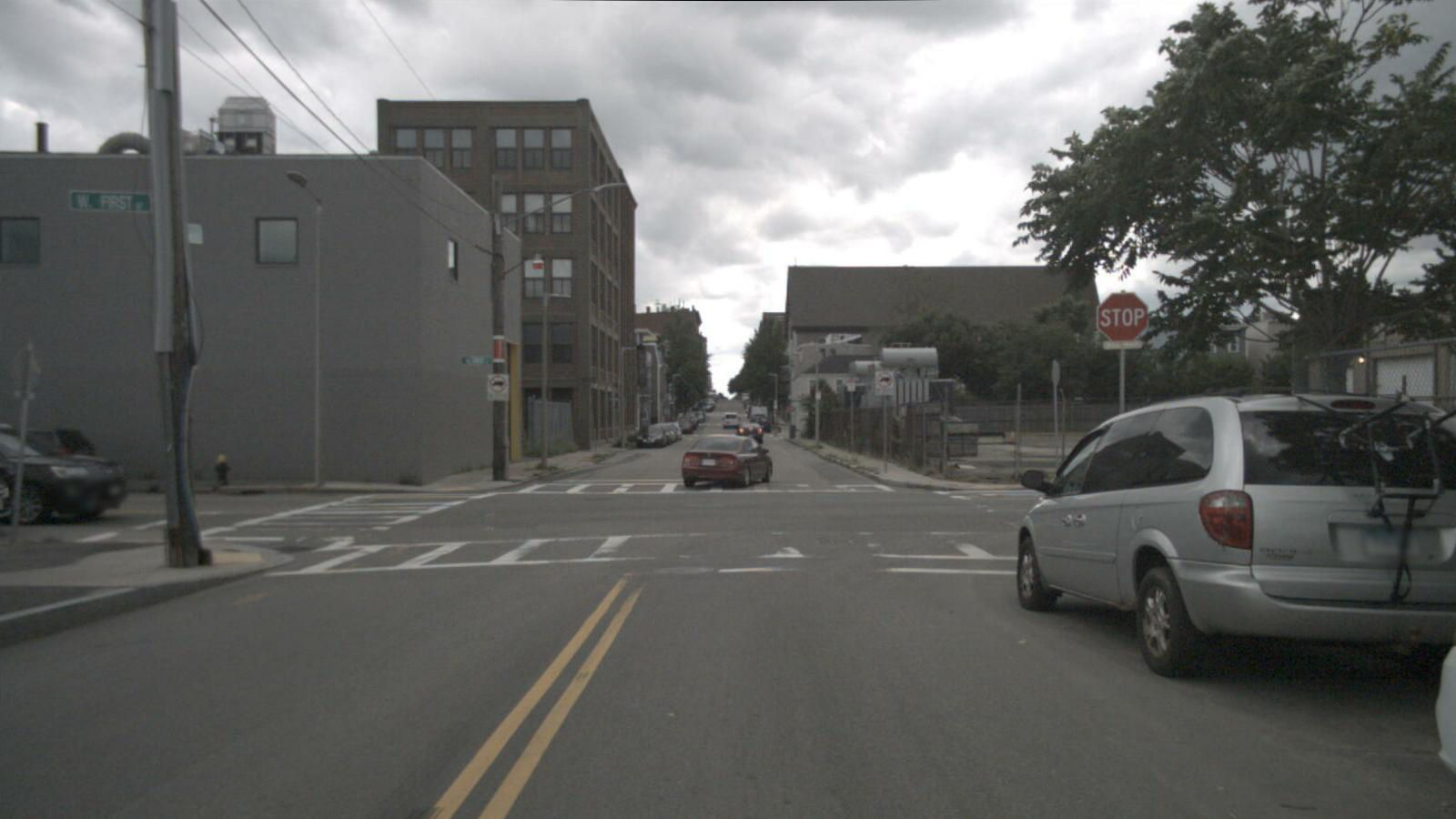}}
    \end{minipage}
    \begin{minipage}{0.26\linewidth}
        \centering
        \subfloat[End-to-end]{\includegraphics[width=\linewidth]{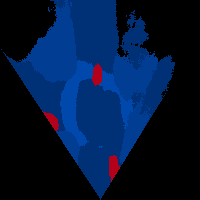}}
    \end{minipage}
    \begin{minipage}{0.26\linewidth}
        \centering
        \subfloat[TaDe (Ours)]{\includegraphics[width=\linewidth]{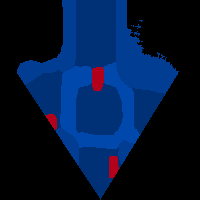}}
    \end{minipage}
    \caption{Qualitative comparison between traditional end-to-end method and our method. (a) shows the perspective view RGB image, (b) and (c) show the BEV segmentation maps predicted without and with the method of task decomposition, respectively.}
    \label{fig:1}
    \vspace{-5pt}
\end{figure}

In the field of autonomous driving, bird's eye view (BEV) perception tasks can obtain adequate semantic information such as road layout, object categories, positions, and scales around vehicles. This kind of information is called BEV representation and can be directly applied to downstream tasks such as layout mapping, action prediction, route planning, and collision avoidance. In this work, we focus on the monocular BEV segmentation task, which is a fundamental task that aims to predict road segments from camera inputs.

Previous BEV segmentation methods~\cite{lu2019monocular,roddick2020predicting,zhou2022cross,zou2023hft,li2022bevformer,saha2022translating,chen2022polar,jiang2023polarformer,peng2023bevsegformer,mani2020monolayout} usually follow an end-to-end pipeline, which adopts a transformer-based architecture to directly predict BEV segmentation maps from the input RGB images. 
However, different from the general segmentation task, the input images and targets in BEV are of distinct views (perspective view \textit{versus} bird's eye view), thus it is hard to train the model by direct point-to-point optimization.
And it is even harder when attempting to achieve both source information perception and target view generation in a single step.  
As shown in Figure~\ref{fig:1} (b), the generated BEV segmentation map of the end-to-end baseline looks distorted and does not conform to the expected regular patterns found in real-world traffic scenes.

To achieve the complex cross-view BEV segmentation, we propose a Task Decomposition (TaDe) method that decompose the traditional end-to-end pipeline into two distinct but goal-oriented stages. In the first stage, we train an autoencoder model to learn a basic understanding of typical BEV scene patterns, enabling it to generate coherent and rational BEV segmentation maps independent from RGB images.
In the second stage, given input RGB images, we only intend to obtain the latent BEV feature, leaving the workload of BEV segmentation map prediction to the pretrained decoder of the first stage.
This two-stage approach, with each stage focusing on only one mission, equips the model with the capacity to handle more intricate and challenging scene perception. As shown in Figure~\ref{fig:1} (c), with the help of task decomposition, the predicted BEV segmentation maps share greater structural similarity with the RGB image.

To be specific, in the first stage, to generate BEV maps that are consistent with physical rules, it is crucial to have a specific expert to comprehend the global traffic layout. For instance, walkways are usually to be continuous rather than discontinuous, and cars are supposed to be shaped as filled rectangles instead of hollow circles. To incorporate this prior knowledge, we employ an autoencoder to learn how to reconstruct/generate the target BEV segmentation maps.
To learn a more intrinsic and robust generating ability, we disturb the autoencoder by corrupting the latent representation with random noise during training and enforce the model perfectly reconstruct the ground truth with only the noised latent.
This reconstruction can locally recognize and predict typical patterns in BEV segmentation maps, thereby ensuring reasonable BEV map generation given imperfect latent representations.

In the second stage, we aim to learn a smooth transition from the perspective RGB input to the BEV semantic latent space. 
Specifically, the model encodes the input RGB images into features and leverages a column-wise transformer to transit them into the BEV latent space. 
The training objective of the second stage is to ensure that the transited RGB representations align seamlessly with the target BEV latent space. Thus we input the BEV label maps to the frozen BEV encoder of Stage I to get the BEV latent feature, and then take this BEV feature as the optimization target for the transition model of Stage II.
Different from existing works, the optimization is in the autoencoder latent space, rather than the traditional pixel segmentation classification space.
During inference, since the representations of both views have been already aligned, the transited RGB feature can directly go through the BEV decoder of Stage I for the final BEV segmentation mask prediction.

Furthermore, to achieve better cross-view alignment, we introduce a coordinate transform step that converts the original BEV segmentation map from Cartesian to polar coordinates. In the transformed BEV map, each column shares the same spatial information with the input monocular RGB images, which also benefits the smooth transition of cross-view features in the second stage. Extensive experiments on two large-scale datasets including nuScenes~\cite{caesar2020nuscenes} and Argoverse~\cite{chang2019argoverse} clearly demonstrate our superiority compared to prior end-to-end works. 

Our contributions can be summarized as follows:

\begin{itemize}
    \item We break down the traditional end-to-end pipeline into two separate stages, with each stage focusing on only one mission (either generation or perception). 
    
    \item We propose a BEV autoencoder to automatically learn typical patterns by reconstructing BEV maps from corrupted latent features, ensuring the decoder prediction is rational and adheres to the real world.
    
    \item We also transform the BEV label map from Cartesian to polar coordinates to learn better alignment across perspective views and bird's eye views. 
\end{itemize}

\section{Related Works}
\subsection{Bird's Eye View (BEV) Perception}
As BEV representations contain rich semantic information and can be directly applied to numerous downstream tasks, BEV perception has received widespread attention in the field of autonomous driving.
However, converting the perspective view captured by cameras to the bird's eye view is an ill-posed problem, which makes it hard to meet the autonomous driving requirements of complex real-world scenarios.
Early works~\cite{philion2020lift,hu2021fiery,huang2021bevdet,zhang2022beverse,park2022time,pan2020cross,li2023bevdepth,liu2023bevfusion} addressed this challenge based on a bottom-up strategy. They first predict the position in the 3D space by estimating the depth of each pixel in the perspective view and then project the 3D position to the BEV space based on the internal and external parameters of the cameras. Nevertheless, despite being intuitive, bottom-up methods depend on the accuracy of depth estimation and have high computational complexity. 
Recent works follow a top-down strategy. Inspired by DETR~\cite{carion2020end}, DETR3D~\cite{wang2022detr3d} detects objects from object queries in 3D space with a transformer. BEVFormer~\cite{li2022bevformer} and BEVFormer v2~\cite{yang2023bevformer} add dense object queries based on DETR3D to achieve map segmentation. PETR~\cite{liu2022petr} and PETRv2~\cite{liu2023petrv2} introduce 3D position encoding to DETR3D and significantly simplify the pipeline.

\subsection{Monocular BEV Segmentation}
Traditional BEV perception methods focus on calculating the corresponding relationship between the coordinates of the perspective view and the bird's eye view by using the internal and external parameters of the cameras. However, due to the geometric distortion and the presence of overlapping parts and gaps between the perspective views, the perception effect is not ideal. In this regard, some works~\cite{bartoccioni2023lara,chen2022efficient,lu2022learning,saha2021enabling,gosala2022bird,gosala2023skyeye} propose to predict the BEV segmentation map of road layout and objects directly in the case of giving a monocular perspective view. 
PYVA~\cite{yang2021projecting} proposes a GAN-based framework and designs a cross-view conversion module to handle significant differences between views. 
VED~\cite{lu2019monocular} employs a variational encoder-decoder network to directly anticipate a semantic occupancy grid from an image. However, the inclusion of a fully connected bottleneck layer in the network results in the loss of significant spatial context and an output that is relatively coarse and incapable of capturing smaller objects like pedestrians. 
CVT~\cite{zhou2022cross} designs a transformer and utilizes the camera's intrinsic and extrinsic calibrations to efficiently understand the surrounding environment.
PON~\cite{roddick2020predicting} adopts a semantic Bayesian occupancy grid framework to estimate BEV segmentation maps directly from monocular perspective views. 
HFT~\cite{zou2023hft} provides a balanced approach that combines structured geometric information from model-based methods with the global context-capturing ability of model-free methods to overcome their respective limitations in image processing. 
DiffBEV~\cite{zou2023diffbev} and DDP~\cite{ji2023ddp} propose an end-to-end framework to generate a more comprehensive BEV representation and denoise noisy samples by applying the diffusion model to BEV perception. 

\subsection{Polar-based Methods}

So far, many works have made significant progress in the BEV segmentation task. However, due to the fact that the perspective view is established in a 3D polar coordinate system, while the bird's eye view is established in a 2D Cartesian coordinate system with the camera as the origin, this gap has led to the previous method's unsatisfactory performance.
Therefore, in order to reduce this gap, some methods choose to convert the ground truth of the bird's eye view in the Cartesian coordinate system to the polar coordinate system, corresponding to the perspective view, in order to better learn the relationship between them.
PolarDETR~\cite{chen2022polar} converts the object representation in DETR3D from the Cartesian coordinate system to the polar coordinate system. 
PolarFormer~\cite{jiang2023polarformer} incorporates a polar alignment module to aggregate rays from multiple cameras and generate a structured polar feature map. 
TIM~\cite{saha2022translating} formulates BEV segmentation map generation from an image as a set of sequence-to-sequence translations and proposes a novel form of end-to-end transformer network to translate between a vertical scanline in the image and rays passing through the camera location in the bird's eye view. 
GitNet~\cite{gong2022gitnet} first learns visibility-aware features and learnable geometry to translate into BEV space and then deforms the pre-aligned coarse BEV features with visibility knowledge by ray-based transformers. 

\section{Method}
We propose TaDe (\textbf{Ta}sk \textbf{De}composition) to break down the traditional end-to-end training pipeline into two simple and distinct stages: (1) With the BEV autoencoder, BEV segmentation maps with 50\% noise are reconstructed to recognize and predict the typical patterns and obtain the BEV latent representations. (2) With the RGB-BEV alignment, the input monocular RGB images are mapped into the BEV latent space via a column-wise transformer to achieve view alignment. During testing, the BEV decoder is employed to generate BEV semantic segmentation maps. In this section, we first introduce the coordinate system transformation of target BEV maps in Section~\ref{sec:3.1}. After that, we elaborate on the BEV autoencoder and column-wise feature mapping in Section~\ref{sec:3.2}. and Section~\ref{sec:3.3}., respectively. 

\subsection{BEV Coordinate System Transformation}
\label{sec:3.1}
\begin{figure}[t]
    \centering
    \includegraphics[width=\linewidth]{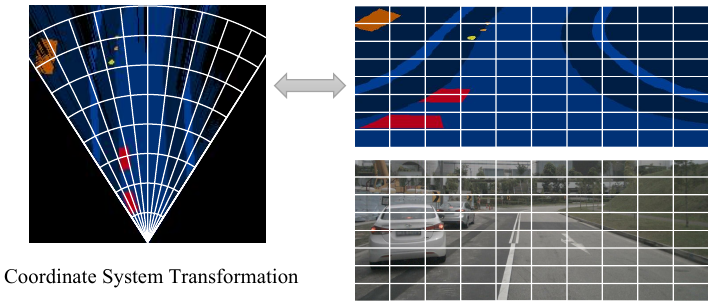}
    \caption{Illustration of the transformation between Cartesian (left) and polar (right) coordinate system for BEV segmentation maps. This transformation can achieve a column-wise correspondence between the BEV segmentation maps and the RGB images.}
    \label{fig:transform}
    \vspace{-5pt}
\end{figure}

\begin{figure*}[t]
    \centering
    \includegraphics[width=\linewidth]{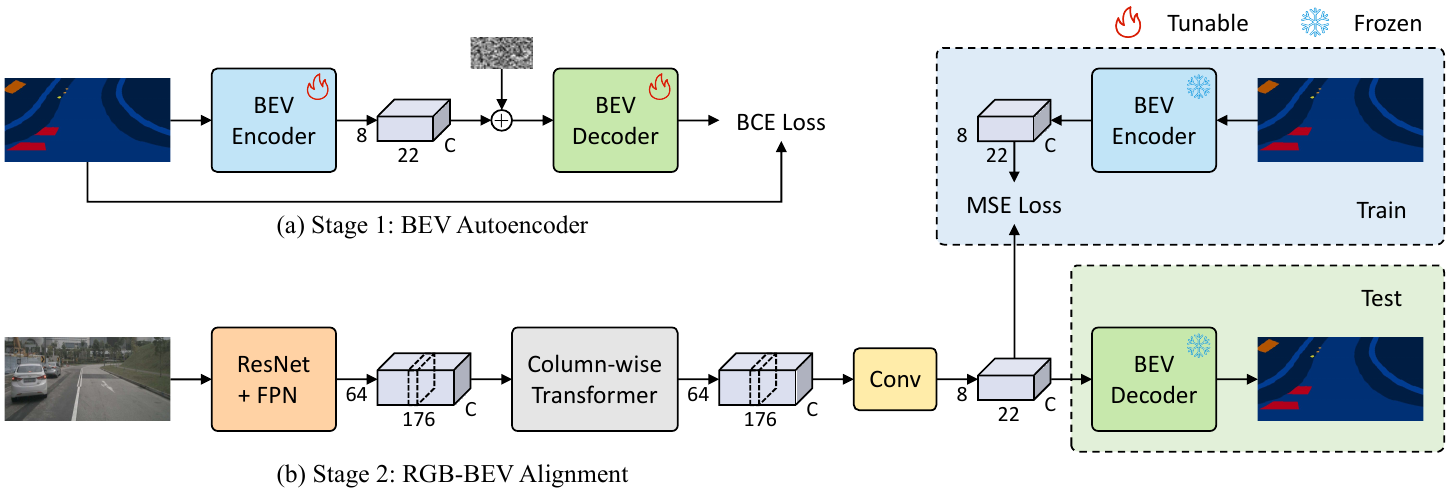}
    \caption{Overview of our two-stage method. (a) In the first stage, a BEV autoencoder is trained by BEV segmentation maps independent from RGB images. (b) In the second stage, the BEV autoencoder is frozen and the RGB-BEV alignment is conducted to match the RGB images to BEV latent representations for decoding.}
    \label{fig:overview}
    \vspace{-5pt}
\end{figure*}

In the monocular BEV semantic segmentation task, the input RGB images are captured from a perspective view in polar coordinates, while the segmented output targets originate from the bird's eye view in Cartesian coordinates. 
Consequently, there exists no pixel-level correspondence between the input RGB images and the target segmentation maps, unlike traditional semantic segmentation tasks, which makes this task extremely challenging.
However, previous polar-based methods~\cite{saha2022translating,gong2022gitnet} have identified a corresponding relationship between the columns of RGB images and the rays in BEV segmentation maps with the camera position as the origin. To establish this correspondence, they conducted resampling from polar to Cartesian coordinates in the feature space. In contrast, to facilitate BEV autoencoder training and achieve column-wise feature alignment, we directly apply coordinate system transformation on the BEV segmentation maps.

To be specific, as illustrated in Figure~\ref{fig:transform}, we employ a resampling method to transform the BEV segmentation maps between the Cartesian and polar coordinate systems to achieve a column-wise spatial correspondence between the BEV segmentation maps and RGB images. For instance, in the Cartesian to polar conversion, for each pixel in the polar map, we calculate the corresponding coordinate in the Cartesian map according to the field of view (FOV) and resample from the nearby pixels, and vice versa. Note that this transform is deterministic and does not require training. During training, we transform the BEV segmentation maps from the Cartesian to the polar coordinate system for reconstruction and encoding. During testing, we transform the predicted BEV segmentation maps from the polar to the Cartesian coordinate system for further quantitative and qualitative validation.

\subsection{BEV Autoencoder}
\label{sec:3.2}

To achieve rational BEV map generation, it is essential to possess a profound understanding of the traffic scene occupancy patterns. In the real world, there is prior knowledge within the traffic BEV segmentation map, where walkways are usually to be continuous rather than discontinuous, and cars are supposed to be shaped as filled rectangles instead of hollow shapes. However, in previous methods, the generated BEV segmentation map does not match real traffic patterns effectively.
To address this issue, we introduce a BEV autoencoder, which is designed to reconstruct the BEV segmentation map in polar coordinates. Notably, we introduce 50\% noise into the BEV latent representation for reconstruction. This addition of noise renders the reconstruction task more challenging, which requires a holistic understanding of the traffic scene patterns and thus ensures the creation of rational BEV maps that align more closely with real-world expectations.

As illustrated in Figure~\ref{fig:overview} (a), the BEV autoencoder consists of a BEV encoder denoted as $E(\cdot)$, and a BEV decoder denoted as $D(\cdot)$. Given a BEV segmentation map $y$ in polar coordinates, the BEV autoencoder is outlined as follows:
\begin{equation}
   z=E(y),
   \label{eq:encoder}
\end{equation}
\vspace{-10pt}
\begin{equation}
   \hat{y}=D(\tilde{z}),
   \label{eq:decoder}
\end{equation}
where $z$ and $\hat{z}$ denotes the BEV latent representation, $\hat{y}$ denotes the reconstructed vector in input space. Note that we obtain $z$ with a low scale of 8 $\times$ 22, which can thus be seen as a latent representation of $y$.

Due to the simplicity of BEV segmentation maps, the criterion is not sufficient to obtain a useful representation or effective decoder. Therefore, we added a large amount of Gaussian noise (\textit{e.g.} 50\%) to the latent feature map $z$, to force the BEV decoder to predict the regular traffic scene patterns from noise. The latent feature with noise is calculated by:
\begin{equation}
   \tilde{z}=\sqrt{1-\eta} z+\sqrt{\eta} \varepsilon,
\end{equation}
where $\eta$ is set to $0.5$ in our experiments and $\varepsilon$ denotes the standard Gaussian distribution, which shares the same dimensionality with $z$.

Ultimately, we train the BEV autoencoder utilizing a weighted BCE loss to reconstruct the input $y$ as accurately as possible, ensuring that the latent feature map $z$ retains as much information from the original input $y$ as possible:
\begin{equation}
   \mathcal L_{BCE}=-w(y\log \hat{y}+(1-y)\log (1-\hat{y})),
   \label{eq:bce_loss}
\end{equation}
where $w$ denotes the training weights of each class. During implementation, the reciprocal of the frequency of occurrence is used as the weight for each class, which enables balanced learning for each type of object.

With the pretrained BEV autoencoder, the BEV latent representation is leveraged for RGV-BEV alignment, while the BEV decoder is utilized for BEV segmentation map generation during testing.

\subsection{RGB-BEV Alignment}
\label{sec:3.3}
The goal of RGB-BEV alignment is to learn a transition from the perspective RGB images to the BEV latent representations. 
As shown in Figure~\ref{fig:overview} (b), the perspective RGB image is taken as the input into the pipeline, where the output is supposed to predict the latent representation $z$.

Following previous works~\cite{roddick2020predicting,saha2022translating}, we first resize the input RGB images from 900 $\times$ 1600 to 396 $\times$ 704, and then crop off the upper part that contains inessential information like the sky and buildings, leaving only the bottom 256 $\times$ 704, in order to better adapt to our model.

At the beginning of the pipeline, we primarily utilize ResNet~\cite{he2016deep} and FPN~\cite{lin2017feature}, denoted as $F(\cdot)$ together, to extract the feature maps $f$ with a lower scale of 64 $\times$ 176 from the preprocessed input RGB image $x$:
\begin{equation}
    f=F(x).
\end{equation}

Despite the fact that we have transformed the BEV segmentation map from Cartesian to polar coordinates to establish spatial correspondence with RGB images from a perspective view of polar coordinates, it is noteworthy that for each column, objects in the RGB image adhere to the characteristic of being larger closer and smaller farther away, whereas objects in the BEV segmentation map maintain the same scale regardless of their distance. To address this, we treat this task as a column-wise sequence-to-sequence translation problem and employ a transformer~\cite{vaswani2017attention} following~\cite{saha2022translating,gong2022gitnet}.
Specifically, we first permute and reshape the feature map from $f^{N\times C\times H\times W}$ to $f^{H\times (N\times W)\times C}$ and treat each column as a sequence input to the transformer~\cite{vaswani2017attention} to produce a feature map $t^{H\times (N\times W)\times C}$, which is then resized back to the scale of $t^{N\times C\times H\times W}$.

For the last module of the pipeline, we apply multiple convolutional layers, denoted as $conv(\cdot)$, to reduce the scale (64 $\times$ 176) of the feature map $t$ obtained from the transformer to match the scale ($8\times 22$) of the feature map $z$,
\begin{equation}
    \hat{z}=conv(t),
\end{equation}
where $\hat{z}$ is the output of the whole pipeline.

In the training phase, we utilize the mean squared error (MSE) loss to facilitate the output of the pipeline to approach the latent feature map $z$ generated by the frozen BEV encoder according to Equation~\ref{eq:encoder}:
\begin{equation}
    \mathcal L_{MSE} = (\hat{z}-z)^2.
    \label{eq:mse_loss}
\end{equation}

During testing, we use the frozen BEV decoder to decode the feature map $\hat{z}$ predicted by the pipeline into the BEV segmentation map according to Equation~\ref{eq:decoder}. Finally, we transform it from the polar to the Cartesian coordinates system following Section~\ref{sec:3.1}.

\begin{table*}[htbp]
 \centering
 \resizebox{\textwidth}{!}{
 \begin{tabular}{cccccccccccccccc}
    \toprule
    \multirow{2}{*}{Method} & \multicolumn{4}{c}{Layout} & \multicolumn{10}{c}{Object} & \multirow{2}{*}{Mean} \\
    \cmidrule(lr){2-5}\cmidrule(lr){6-15}
             & Drivable & Crossing & Walkway & Carpark & Bus & Bike & Car & C.V. & Motor. & Trailer & Truck & Ped. & Cone & Barrier\\
    \midrule
    IPM~\cite{roddick2020predicting}    & 40.1 & - & 14.0 & - & 3.0 & 0.2 & 4.9 & - & 0.8 & - & - & 0.6 & - & - & -\\
    Depth Unpr.~\cite{roddick2020predicting}    & 27.1 & - & 14.1 & - & 6.7 & 1.3 & 11.3 & - & 2.8 & - & - & 2.2 & - & - & -\\
    VED~\cite{lu2019monocular}    & 54.7 & 12.0 & 20.7 & 13.5 & 0.0 & 0.0 & 8.8 & 0.0 & 0.0 & 7.4 & 0.2 & 0.0 & 0.0 & 4.0 & 8.7\\
    VPN~\cite{pan2020cross}    & 58.0 & 27.3 & 29.4 & 12.9 & 20.0 & 4.4 & 25.5 & 4.9 & 5.6 & 16.6 & 17.3 & 7.1 & 4.6 & 10.8 & 17.5\\
    Sim2real~\cite{reiher2020sim2real} & 60.5 & 27.1 & 19.2 & 18.3 & 6.9 & 3.8 & 7.1 & 0.3 & 4.5 & 3.2 & 4.7 & 1.8 & 4.2 & 12.1 & 12.4\\
    OFT~\cite{roddick2018orthographic}    & 62.4 & 30.9 & 34.5 & 23.5 & 23.2 & 4.6 & 34.7 & 3.7 & 6.6 & 18.2 & 17.4 & 1.2 & 1.1 & 12.9 & 19.6\\
    PON~\cite{roddick2020predicting}    & 60.4 & 28.0 & 31.0 & 18.4 & 20.8 & 9.4 & 24.7 & \textbf{12.3} & 7.0 & 16.6 & 16.8 & 8.2 & 5.7 & 8.1 & 19.1\\
    EPOSH~\cite{dwivedi2021bird}  & 61.1 & 33.5 & 37.8 & 25.4 & 31.8 & 6.7 & 37.8 & 2.7 & 10.5 & 14.2 & 20.4 & 5.9 & 7.6 & 13.4 & 22.1\\
    DiffBEV~\cite{zou2023diffbev}& 65.4 & 41.3 & 41.1 & 28.4 & 33.7 & 13.2 & 38.9 & 8.4 & 14.4 & 21.1 & 23.1 & 9.6 & 7.5 & 16.7 & 25.9\\
    GitNet~\cite{gong2022gitnet} & 65.1 & \textbf{41.6} & 42.1 & \textbf{31.9} & 35.4 & 13.8 & \textbf{43.4} & 9.7 & 15.0 & 22.5 & 25.5 & \textbf{14.1} & 11.6 & 18.6 & 27.9\\
    \midrule
    \textbf{TaDe (Ours)} & \textbf{65.9} & 40.9 & \textbf{42.3} & 30.7 & \textbf{38.5} & \textbf{14.2} & 42.8 & 11.4 & \textbf{18.2} & \textbf{26.3} & \textbf{26.9} & 14.0 & \textbf{15.0} & \textbf{21.4} & \textbf{29.2}\\
    \bottomrule
 \end{tabular}
 }
 \caption{Quantitative results of IoU (\%) on the nuScenes~\cite{caesar2020nuscenes} validation set. ``Mean'' indicates IoU across all classes. ``Crossing'',  ``C.V.'', ``Motor.'', ``Ped.'', and ``Cone'' denote pedestrian crossing, construction vehicle, motorcycle, pedestrian, and traffic cone, respectively. IPM and Depth Unpr. are proposed in PON~\cite{roddick2020predicting}. }
 \label{tab:nuscenes}
\end{table*}

\begin{table*}[htbp] \small
 \centering
 \scalebox{0.9}{
 \begin{tabular}{cccccccccc}
    \toprule
    Method & Drivable & Vehicle & Pedestrian & Large Vehicle & Bicycle & Bus & Trailer & Motorcycle & Mean\\
    \midrule
    IPM~\cite{roddick2020predicting}    & 43.7 & 7.5 & 1.5 & - & 0.4 & 7.4 & - & 0.8 & -\\
    Depth Unpr.~\cite{roddick2020predicting}    & 33.0 & 12.7 & 3.3 & - & 1.1 & 20.6 & - & 1.6 & -\\
    VED~\cite{lu2019monocular}    & 62.9 & 14.0 & 1.0 & 3.9 & 0.0 & 12.3 & 1.3 & 0.0 & 11.9\\
    VPN~\cite{pan2020cross}    & 64.9 & 23.9 & 6.2 & 9.7 & 0.9 & 3.0 & 0.4 & 1.9 & 13.9\\
    PON~\cite{roddick2020predicting}    & 65.4 & 31.4 & 7.4 & 11.1 & 3.6 & 11.0 & 0.7 & 5.7 & 17.0\\
    
    \midrule
    \textbf{TaDe (Ours)} & \textbf{68.3} & \textbf{34.5} & \textbf{9.3} & \textbf{14.7} & \textbf{4.4} & \textbf{37.8} & \textbf{3.1} & \textbf{6.4} & \textbf{22.3} \\
    \bottomrule
 \end{tabular}}
 \caption{Quantitative results of IoU (\%) on the Argoverse~\cite{chang2019argoverse} validation set.}
 \label{tab:argoverse}
 \vspace{-5pt}
\end{table*}

Although we can decode the aligned latent feature map $\hat{z}$ predicted from the pipeline with the frozen BEV decoder into the BEV segmentation map and achieve good performance, we observe that the decoder has only been exposed to samples of BEV segmentation maps during its training process and thus struggles to adequately adapt to the data distribution of input RGB images.
To address this issue, we introduce a fine-tuning strategy that enables the decoder to better adapt to the data distributions of both input RGB images and latent features, thus establishing a connection between the data distribution of BEV segmentation maps and that of input RGB images.
To achieve this, we freeze the parameters of the pipeline and only apply fine-tuning to the BEV decoder using the BCE loss in Equation~\ref{eq:bce_loss}.

\section{Experiments}
\subsection{Experimental Setup}

\textbf{Datasets.}
The nuScenes dataset~\cite{caesar2020nuscenes} comprises 1,000 scenes captured by six cameras in Boston and Singapore. Each scene spans approximately 20 seconds. For semantic mapping, nuScenes provides detailed vectorized maps covering 11 semantic classes. In terms of dynamic objects, the dataset includes comprehensive annotations of 3D bounding boxes for 23 classes, complemented by 8 unique attributes. 
The Argoverse dataset~\cite{chang2019argoverse} includes 65 training and 24 validation sequences recorded in Miami and Pittsburgh, utilizing seven surround-view cameras. Similar to the nuScenes dataset, Argoverse offers 3D object annotations across 15 categories and detailed semantic map data that encompasses road masks, lane geometry, and ground elevation. For a fair comparison, we use the same ground truth generation process, layout and object classes, and training/validation splits outlined in PON~\cite{roddick2020predicting}.

\textbf{Evaluation Metric.}
We present our results using the Intersection-over-Union (IoU) metric across background classes. Additionally, we compute the class-averaged mean IoU to assess overall performance. We ignore invisible grid cells during the evaluation process, as a LiDAR ray that cannot penetrate them will render them invisible in the image. Considering the possibility of class overlap, such as a car driving in the drivable area, we perform an independent binary segmentation evaluation for each category. We then determine the best performance by selecting the maximum IoU from various thresholds.

\textbf{Implementation Details.}
Referring to PON~\cite{roddick2020predicting}, we perform segmentation in a $[0, 50]\times[-25, 25]$ square meters region around the ego car with a resolution of 0.25 meters per pixel, resulting in a final map size of $200\times200$. 
For each stage and fine-tuning, we conduct training for 50 epochs. 
We employ the Lion optimizer~\cite{chen2024symbolic}, set the initial learning rates at $5\times10^{-4}$, $2\times10^{-5}$, and $2\times10^{-4}$ for the first, second, and fine-tuning stages, respectively. Additionally, we use a weight decay factor of 0.01, complemented by warm-up and cosine learning rate decay strategies.

\subsection{Quantitative Comparison}
Although our method predicts based on monocular RGB images without exploring the relationships between multiple cameras, we conducted experiments on all available surrounding cameras in nuScenes~\cite{caesar2020nuscenes} and Argoverse~\cite{chang2019argoverse} and showed the average results on their validation set following most previous works~\cite{roddick2020predicting,gong2022gitnet} in Table~\ref{tab:nuscenes} and Table~\ref{tab:argoverse}. 

\textbf{nuScenes.} As shown in Table~\ref{tab:nuscenes}, we observe that our method surpasses IPM-based approaches like IPM~\cite{roddick2020predicting} and Sim2real~\cite{reiher2020sim2real} by at least 16.8 mIoU, without relying on the assumption of a perfectly flat ground plane. Our method also outperforms bottleneck-based approaches, including VED~\cite{lu2019monocular}, VPN~\cite{pan2020cross}, and PON~\cite{roddick2020predicting}, because our BEV decoder can effectively decode latent features.   
Compared to depth-based approaches like Depth Unprojection-based (DepthUnpr.)~\cite{roddick2020predicting}, OFT~\cite{roddick2018orthographic}, EPOSH~\cite{dwivedi2021bird}, and DiffBEV~\cite{zou2023diffbev}, our method outperforms them by at least 3.3 mIoU, without depending on depth estimation results. 
Furthermore, our method achieves the highest performance of 29.2 on mIoU and surpasses the state-of-the-art method GitNet~\cite{gong2022gitnet} in 10 classes.
The above experimental results demonstrate that our method has achieved a higher level of accuracy, particularly in the segmentation of objects such as motorcycles, trailers, and traffic cones.

\begin{table}[t]
 \centering
 \scalebox{0.85}{
 \begin{tabular}{ccccc}
    \toprule
    Method & Params$\downarrow$ & FLOPs$\downarrow$ & FPS$\uparrow$ & GPU Hours$\downarrow$\\
    \midrule
    PON~\cite{roddick2020predicting}    & \textbf{38.6M} & 60.9G & 43.8 & 100\\
    TIM~\cite{saha2022translating}    & 45.7M & 268.3G & 18.1 & 160\\
    \textbf{TaDe (Ours)} & 41.5M & \textbf{48.5G} & \textbf{51.4} & \textbf{8 + 48 + 24 = 80}\\
    \bottomrule
 \end{tabular}}
 \caption{Comparison of computational overhead on nuScenes~\cite{caesar2020nuscenes}. We report the GPU hours of BEV autoencoder, RGB-BEV alignment, and fine-tuning of our method respectively.}
 \vspace{-5pt}
 \label{tab:overhead}
\end{table}

\begin{figure*}
    \centering
    \includegraphics[width=\textwidth]{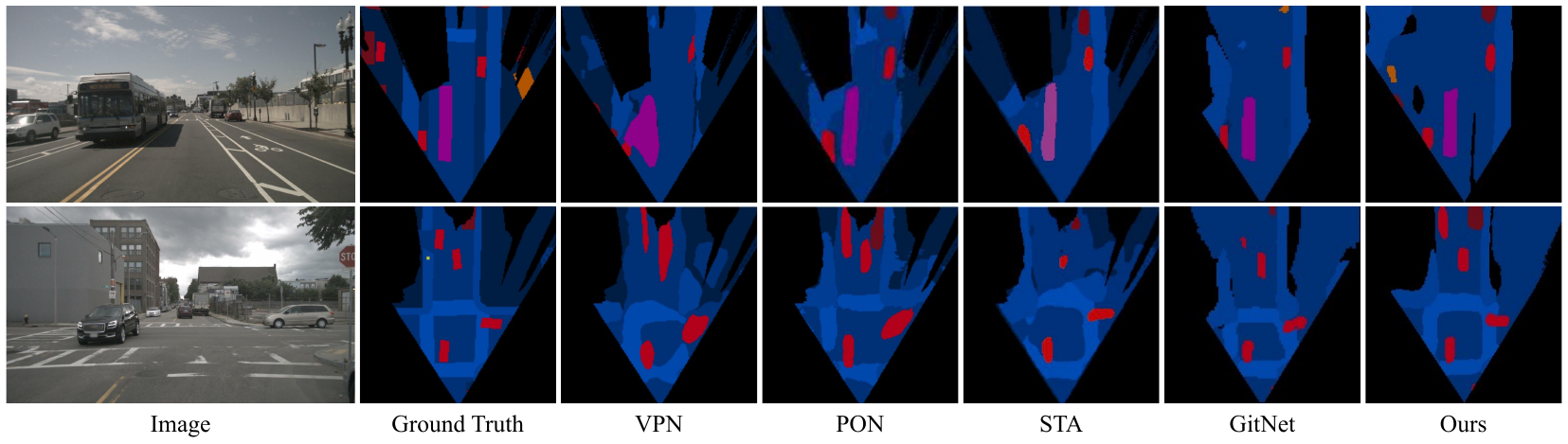}
    \caption{Qualitative results on nuScenes~\cite{caesar2020nuscenes}. We compare with other methods following the color scheme used in PON~\cite{roddick2020predicting}.}
    \vspace{-5pt}
    \label{fig:nuscene visualize}
\end{figure*}

\textbf{Argoverse.} As shown in Table~\ref{tab:argoverse}, we observe that our method achieves the highest performance in terms of mIoU and consistently exhibits improvement across all classes compared with PON~\cite{roddick2020predicting}.
This observation demonstrates the superiority and robustness of our method across diverse datasets.

\textbf{Computational Overhead.}
We compare the computational overhead with two open-source methods, considering the following four metrics: parameters, FLOPs, FPS, and GPU hours, as presented in Table~\ref{tab:overhead}. GPU hours represent the estimated GPU (NVIDIA GeForce RTX 4090) hours required for training to converge. Notably, in inference, our method achieves the best FLOPs and FPS. In terms of training, despite three stages, our method consumes 80 GPU hours in total, while TIM~\cite{saha2022translating} requires 160 GPU hours. In summary, compared with state-of-the-art methods, our method achieves the best results and maintains low computational overhead thanks to not using multi-scale features.

\begin{figure}[t]
    \centering
    \begin{minipage}{0.496\linewidth}
        \centering
        \subfloat[Quantitative Comparison]{\includegraphics[width=\linewidth]{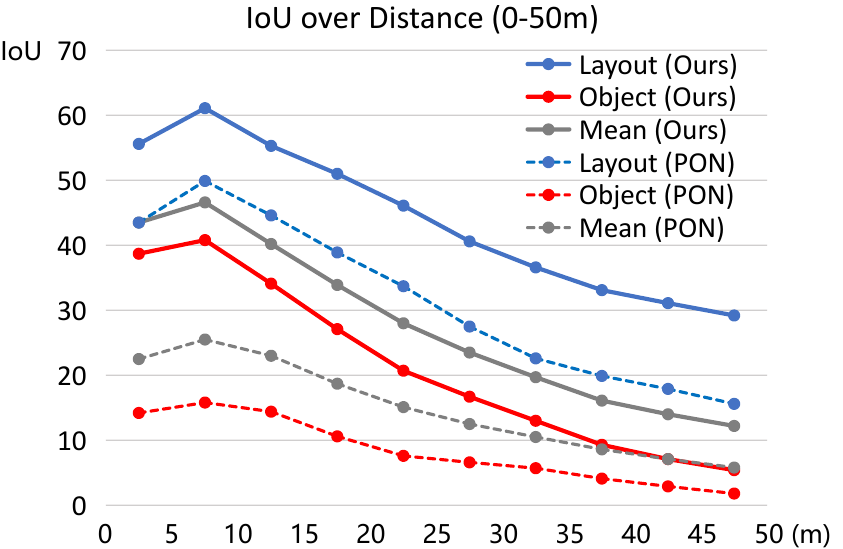}}
    \end{minipage}
    \begin{minipage}{0.48\linewidth}
        \centering
        \subfloat[Percentage Improvement]{\includegraphics[width=\linewidth]{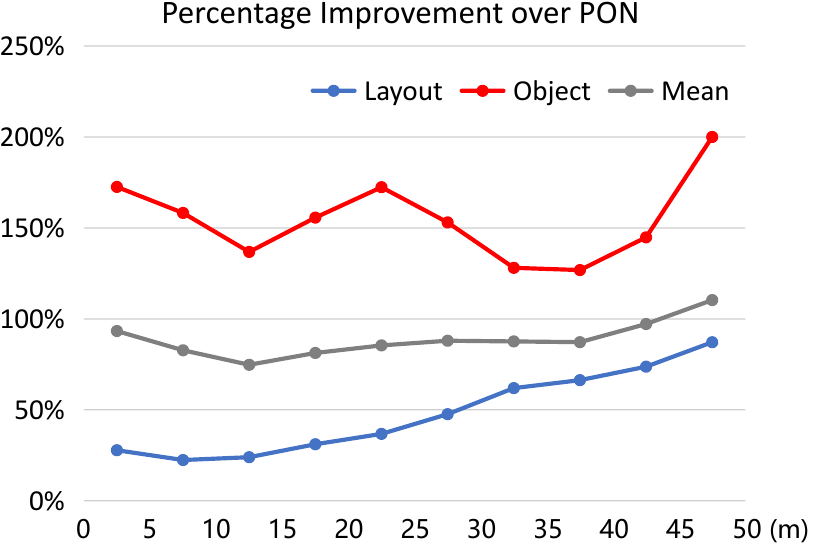}}
    \end{minipage}
    \caption{Trend demonstration and comparison with PON~\cite{roddick2020predicting} of IoU over distance (0-50m) on nuScenes~\cite{caesar2020nuscenes}.}
    \label{fig:distance}
    \vspace{-5pt}
\end{figure}

\textbf{IoU over Distance.}
We further discuss the experimental results of IoU over distance. We divide the BEV maps into 5-meter intervals along the distance from the camera and report the mIoU for layouts and objects at various distances in Figure~\ref{fig:distance}. The experimental results met our expectations, with IoU showing a decreasing trend as the distance between the target and the camera increases. Compared to PON~\cite{roddick2020predicting}, we achieve significant and consistent improvements across distances. Notably, for layouts such as drivable areas, we obtain even more absolute performance gains at larger distances. As for small objects like pedestrians and bikes, we observe a relative performance improvement of 200\% (5.4 vs 1.8) at 45-50 meters, while the relative improvement at 0-5 meters is 173\% (38.7 vs 14.2). In summary, our method shows its superiority, especially at large distances for both layouts and objects.

\subsection{Ablation Study}
In this section, we explore the effectiveness of each proposed component through an ablation study on nuScenes~\cite{caesar2020nuscenes}.
The experimental groups and results are shown in Table~\ref{tab:ablation}. 

\begin{figure*}[t]
    \centering
    \includegraphics[width=0.85\linewidth]{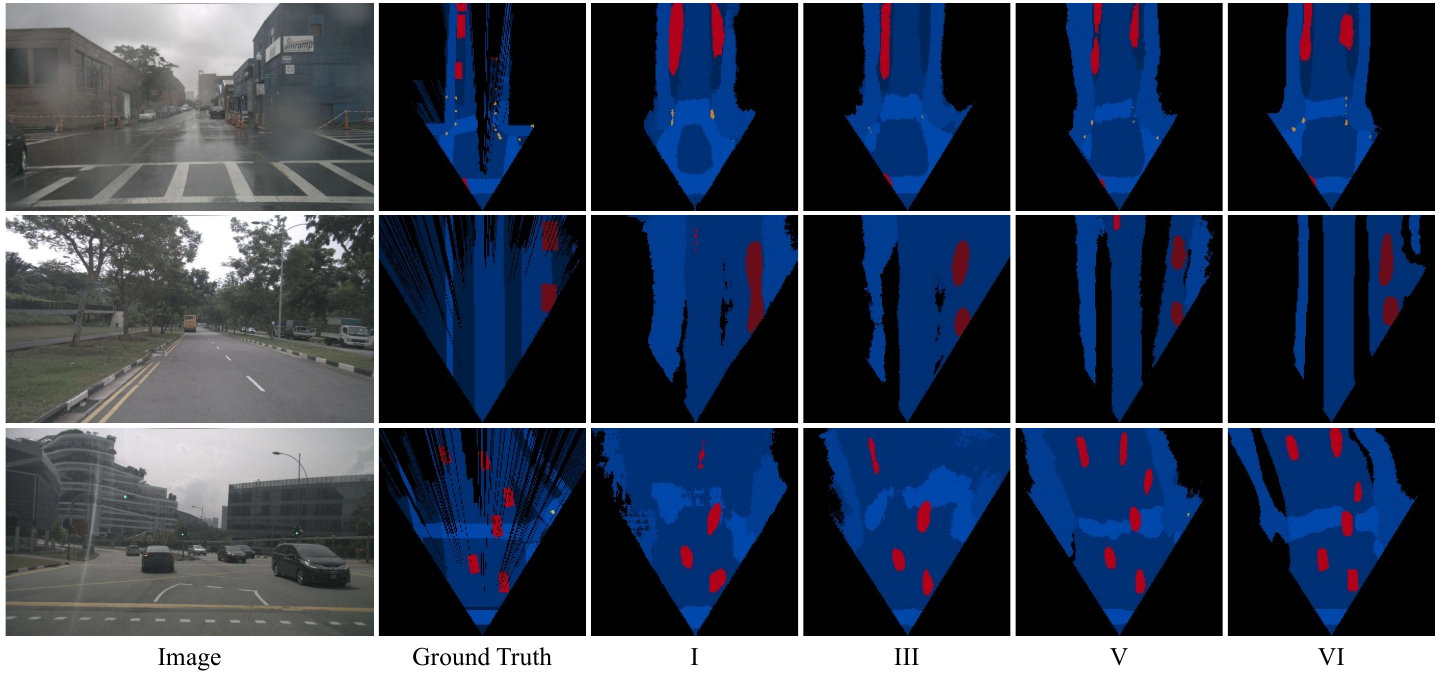}
    \caption{Qualitative results of ablations on nuScenes~\cite{caesar2020nuscenes}. I, III, V, and VI are corresponded to the relevant ablations in Table~\ref{tab:ablation}.}
    \label{fig:abalation}
    \vspace{-5pt}
\end{figure*}

\begin{table}[tbp] 
\centering
\scalebox{0.69}{
\begin{tabular}{ccccccccc} 
    \toprule
     \multirow{2}{*}{Group} & \multicolumn{4}{c}{Components} & \multicolumn{4}{c}{IoU}\\
    \cmidrule(lr){2-5}\cmidrule(lr){6-9}
             &CST& TD & CWT & FT & Layout & Large Obj. & Small Obj. & Mean\\
    \midrule
    \uppercase\expandafter{\romannumeral1} &  & & &  & 35.0 &  22.0 & 10.4 & 21.6 \\
    \uppercase\expandafter{\romannumeral2} &  & \checkmark & &  & 41.9 & 22.9 & 11.7 & 24.4 \\
    \uppercase\expandafter{\romannumeral3} & \checkmark & & &  & 41.5 & 25.6 & 13.8 &25.9 \\
    \uppercase\expandafter{\romannumeral4} & \checkmark &  \checkmark & & & 44.1 & 27.5 & 13.9 &27.4 \\
    \uppercase\expandafter{\romannumeral5} & \checkmark & & \checkmark &  &  41.2 & 26.4 & 15.5 &26.7 \\
    \uppercase\expandafter{\romannumeral6} & \checkmark &  \checkmark & \checkmark && 44.8 & 28.6 & 15.9 &28.7 \\
    \uppercase\expandafter{\romannumeral7} & \checkmark &  \checkmark & \checkmark & \checkmark &\textbf{45.0} & \textbf{29.2} & \textbf{16.6} & \textbf{29.2} \\
    \bottomrule
 \end{tabular}}
 \caption{Ablations of different key components on nuScenes~\cite{caesar2020nuscenes}. CST, TD, CWT, and FT denote coordinate system transformation, task decomposition, column-wise transformer, and fine-tuning, respectively. Large objects include bus, car, construction vehicle, trailer, and truck. Small objects include barrier, bike, motorcycle, pedestrian, and traffic cone.}
 \label{tab:ablation}
 \vspace{-5pt}
\end{table}

\textbf{Coordinate System Transformation.} 
In Table~\ref{tab:ablation}, Groups I and III, compared with Groups II and IV, are designated as control groups. Following the adoption of the BEV segmentation map in polar coordinates, there is an overall increase in mIoU by 4.3 and 3, respectively. This improvement can be attributed to the column-wise correspondence established between the BEV segmentation map in polar coordinates and the perspective image, enabling the model to make more accurate predictions. Additionally, Group III exhibits a notable improvement, showing increases of 6.5 in layout and 3.6 in small objects compared to Group I.

\textbf{Task Decomposition.}
We deploy ablations to demonstrate the effectiveness of task decomposition. For Groups I, III, and V in Table~\ref{tab:ablation}, we train the pipeline and the BEV decoder simultaneously with the BCE loss in an end-to-end paradigm. Comparing Group V and Group VI, we observe that task decomposition brings a significant improvement to layouts (41.2 vs 44.8) and large objects (26.4 vs 28.6) which commonly appearing typical patterns. Similar improvements could be observed in the comparison between Group III and Group IV, especially between Group I and Group II (6.9 IoU improvement to layouts). The above ablations demonstrate that the BEV autoencoder successfully learned prior patterns of traffic scenes to decode a more rational BEV segmentation map.

\textbf{Column-wise Transformer.}
In RGB images, objects exhibit the characteristic of appearing larger when closer and smaller when farther away, while in BEV segmentation maps, objects maintain the same scale regardless of their distance. Motivated by TIM~\cite{saha2022translating}, we adopt a column-wise transformer to translate each column of the feature map to match the characteristics between the perspective view and the bird's eye view. As shown in Table~\ref{tab:ablation}, when compared with Group IV and Group VI, the column-wise transformer improves the IoU of small objects significantly (13.9 vs 15.9) while also improving the IoU of large objects (27.5 vs 28.6). Similar improvements are observed in the comparison between Group III and Group V. These experimental results prove that the column-wise transformer could better capture the information of objects.

\textbf{Fine-tuning.} 
Although Stage 1 and Stage 2 for training are sufficient to achieve good results, the BEV decoder was trained independently from RGB images so that could not decode typical patterns in RGB images to a detailed BEV segmentation map. Thus we freeze the pipeline as a pretrained RGB image feature extractor and fine-tune the BEV decoder in an end-to-end manner to establish connections between RGB images and BEV segmentation maps. Compared to Group VI in Table~\ref{tab:ablation}, Group VII achieves better results across all classes, especially on objects that occupy small regions (0.7 IoU improvement).

\subsection{Qualitative Comparison}

Figure~\ref{fig:nuscene visualize} shows qualitative results on nuScenes~\cite{caesar2020nuscenes}, comparing our predicted segmentation maps with other methods and the ground truth. Our segmentation maps appear more precise and rational, benefiting from our task decomposition into two stages. Specifically, our model demonstrates a more coherent road layout, pedestrian crossing, and walkway layout, thanks to prior knowledge learned in advance. For dynamic objects, our model effectively leverages the polar coordinate system to predict nearby and distant objects with high accuracy, particularly excelling in determining the scale of cars and buses.

Figure~\ref{fig:abalation} provides further qualitative results of our ablation experiments. Groups III, V, and VI in the first row display more coherent and accurate road layouts due to the introduction of the coordinate system transformation. Group V in the second row successfully predicts distant cars, thanks to the effectiveness of the column-wise transformer in translating each column. In the third row of Figure~\ref{fig:abalation}, Group VI demonstrates that the BEV autoencoder has successfully learned prior patterns about layouts and objects, accurately predicting the size of cars and the layout of pedestrian crossings compared to other experiments.

\section{Conclusion}
In this paper, we leverage a clearer and more direct optimization target, by decomposing the end-to-end pipeline into two separate stages, with each stage focusing on only one mission (either generation or perception). Specifically, in the first stage, we propose a BEV autoencoder to reconstruct the BEV segmentation map with corrupted noise, where the decoder is forced to learn typical BEV traffic scene patterns. In the second stage, a mapping from input RGB images and BEV latent representations is learned. Furthermore, we transform the BEV segmentation map from a Cartesian coordinate system to a polar one, thus enhancing its alignment with RGB input. 
Notably, our method abstains from utilizing multi-scale features or camera intrinsic parameters for depth estimation, resulting in a significantly lower computational overhead. Additionally, our method consistently achieves outstanding prediction accuracy across varying distances from the camera.
Extensive experiments on two datasets demonstrate our effectiveness. 

\section*{Acknowledgement} 
This work was partially supported by the National Natural Science Foundation of China under Grant 62372341 and sponsored by CCF-DiDi GAIA Collaborative Research Funds for Young Scholars.

{\small
\bibliographystyle{ieee_fullname}
\bibliography{main}
}

\end{document}